# Knee Motion Generation Method for Transfemoral Prosthesis based on Kinematic Synergy and Inertial Motion

Hiroshi Sano, Takahiro Wada, *Member, IEEE*

*Abstract*— Previous research has shown that the effective use of inertial motion (i.e., less or no torque input at the knee joint) plays an important role in achieving a smooth gait of transfemoral prostheses in the swing phase. In our previous research, a method for generating a timed knee trajectory close to able-bodied individuals, which leads to sufficient clearance between the foot and the floor and the knee extension, was proposed using the inertial motion. Limb motions are known to correlate with each other during walking. This phenomenon is called kinematic synergy. In the present study, we measure gaits in level walking of able-bodied individuals with a wide range of walking velocities. We show that this kinematic synergy also exists between the motions of the intact limbs and those of the knee as determined by the inertial motion technique. We then propose a new method for generating the motion of the knee joint using its inertial motion close to the able-bodied individuals in mid-swing based on its kinematic synergy, such that the method can adapt to the changes in the motion velocity. The numerical simulation results show that the proposed method achieves prosthetic walking similar to that of able-bodied individuals with a wide range of constant walking velocities and termination of walking from steady-state walking. Further investigations have found that a kinematic synergy also exists at the start of walking. Overall, our method successfully achieves knee motion generation from the initiation of walking through steady-state walking with different velocities until termination of walking.

## I. INTRODUCTION

RECENT advances in microprocessor-controlled transfemoral prostheses [1-4] have successfully improved the safety of walking with a prosthesis, as well as gait symmetry by introducing knee joint control with variable damping, in which the knee joint does not generate positive power. Muscles are known to perform a certain amount of positive work during level walking, and using a passive prosthesis increases the prosthesis user's muscle load [5]. In addition, there are cases when an active prosthesis, which is defined as the prosthesis that generates positive power in the prosthetic joint using any actuator, is advantageous, such as when ascending a slope. Therefore, several active transfemoral prostheses, in which actuators are used, have been developed [6-14].

Impedance control, in which the behavior of a closed-loop system against the external force is controlled, has been widely used as the low-level controller of the active transfemoral prostheses. For example, the impedance control method, in which the mechanical impedance parameters and their equilibrium point were controlled for each of the predetermined finite states of the walking cycle, was developed in previous studies [6-8] to achieve a complete walking cycle. In impedance control, the knee motion is indirectly generated as the consequence of an external force being exerted, including hip moment and other disturbances. Thus, precisely specifying the resultant knee motion is difficult, especially when there is a disturbance or a fluctuation in the motion of the thigh.

Tracking control, in which a desired motion pattern of a joint is predefined, and the joint is controlled to track it, is also used for active prostheses. For example, an optimal control that minimizes the actuator torque and the tracking error has been investigated [9,10], in which the knee joint of an able-bodied individual measured in advance by fixing the ankle movement with an ankle orthosis is used as the desired motion. Such a position-based tracking control approach that can directly specify the desired knee motions, such as those of able-bodied individuals, is promising to achieve safer walking in the swing phase and keep sufficient clearance between the foot and the floor as well as the knee extension just before the heel contact.

The desired position in the tracking control is often described by the function of time. For example, the knee motion of able-bodied individuals measured in advance is used [9,10]. Echo control, in which the joint motion of the intact side in the current step is used as the desired motion in the next step of the prosthetic side, is also proposed [15]. The use of such a timed trajectory or an "echoing" trajectory is not robust to the changes in the walking velocity or the sudden changes in the walking mode, such as stopping or changes in direction. We focus herein on the fact that many limbs participating in skillful human motions, such as walking, are not independently actuated, but instead work cooperatively. The corresponding kinematic aspect is referred to as kinematic synergy or intersegmental coordination [16-18]. From this fact, we conceived that the desired knee motion can be described by the function of the state of the intact joint at each moment.

This work was supported in part by a JSPS KAKENHI Grant-in-Aid for Scientific Research (B) (grant number 25289065).
H. Sano was with the Graduate School of Information Science and Engineering, Ritsumeikan University, Shiga 525-8577, Japan (E-mail: hiroshisano@hr.ci.ritsumei.ac.jp).

T. Wada is with the College of Information Science and Engineering, Ritsumeikan University, Kusatsu, Shiga 525-8577, Japan (Phone & fax: +81-77-561-2798; E-mail: twada@fc.ritsumei.ac.jp).

Several studies indicate that the inertial motion or dynamics of the entire body is leveraged in the swing phase when the users of the transfemoral prosthesis feel to walk easily [19,20]. Thus, it is hypothesized that a control method, in which the desired motion leverages the inertial motion, would increase the ease of walking. Furthermore, utilizing the inertial motion could increase energy efficiency, which could reduce the size of the actuator and the battery. Using the inertial motion also leads to the decrease of discomfort in walking because the inertial parameters are shown to affect the swing phase motion [19,21,22]. Light-weight prosthesis also leads to less energy consumption of the user [22]. Therefore, it is conceived that generating a knee motion close to able-bodied individuals leveraging the inertial motion can achieve active prostheses with safety and ease of walking. However, note that the mass of the transfemoral prostheses is different from the biological counterparts of able-bodied individuals (e.g., most of the passive transfemoral prostheses are lighter than the biological counterparts of able-bodied individuals). Their inertial motions can be different from each other even with the same initial condition. Therefore, we developed a motion generation method of prosthetic knee that fits with the able-bodied individuals in mid-swing [23]. However, this method could not be applied to situations where there are changes in the walking velocity because the proposed method generates timed trajectories.

Therefore, the first purpose of the current study is to examine whether the kinematic synergy exists even during walking with a transfemoral prosthesis using the inertial motion in the mid-swing, which is similar to that of an intact knee. The second purpose is to exploit this synergy to generate the motion of the prosthetic knee joint that can adapt to the walking velocity changes.

First, the kinematic synergy between the intact limbs and the prosthetic knee employing the inertial motion is investigated. A method for generating the motion of the prosthetic knee joint is proposed based on an analysis of the kinematic synergy to adapt to the walking velocity changes. We found that the walking velocity could be estimated well from the initial motion. Hence, in the proposed method, the coefficients used to generate the knee motion from the biological joints are changed depending on the initial motion at each toe-off. Finally, numerical simulations are performed to evaluate the effectiveness of the proposed method.

## II. KINEMATIC SYNERGY ANALYSIS OF GAIT WITH TRANSFEMORAL PROSTHESIS USING INERTIAL MOTION

### A. Level Walking Experiment by Able-bodied Individuals

*1) Experimental apparatus*
A straight and flat path with a length of 7.5 m was used. A total of 15 reflective markers were attached to the greater trochanter, knees, ankles, heels, fifth metatarsals, and toes of both feet as well as the shoulders and the back of the subjects to measure the positions of the lower extremities and the orientation of the trunk. The three-dimensional positions of the markers were measured using a motion-capture system (motion analysis) with ten cameras at 100 Hz.

*2) Participants and procedure*
Six males, who all provided informed consent and did not have any neuromuscular disorders or functional limitations in their lower extremities, participated in the experiments. Their mean (SD) age was 22.2 (1.1) years. Their mean (SD) body height and weight were 1.74 (0.05) m and 65.9 (9.2) kg, respectively.

The participants were instructed to walk along the predetermined straight path from the start point. Their cadence was determined using a metronome. No instructions regarding the stride were provided. A total of 13 cadences (70–130 bpm in 5 bpm intervals) were determined. Five walking trials per cadence (i.e., a total of 65 trials) were conducted for each subject after some practice walks.

### B. Link Segment Model

The human posture in the swing phase of the prosthesis side was modeled as shown in Fig. 1 [23]. Link 1 denotes the stance leg, including the upper and lower legs, while ignoring the knee joint. The foot joint was fixed on the floor. Link 2 denotes the thigh and the socket of the prosthesis, while link 3 represents the lower leg and the foot part of the prosthesis. Link 0 denotes the trunk and the upper limbs and is assumed to be vertical during the swing phase.

$$H(q)\ddot{q} + \frac{1}{2}\dot{H}(q)\dot{q} + S(q,\dot{q})\dot{q} + g(q) = \tau ,\qquad(1)$$

where $q := [q_{12}^T, q_3]^T = [q_1, q_2, q_3]^T$ and $\tau := [\tau_1, \tau_2, \tau_3]^T$ denote the joint angle vector and the joint torque vector, respectively [24]. $H(q) := [h_{ij}(q)]$ represents the inertia matrix, and $S(q,\dot{q})$ stands for a skew-symmetric matrix representing the centrifugal and Coriolis forces. Vector $g(q)$ denotes the gravity term.

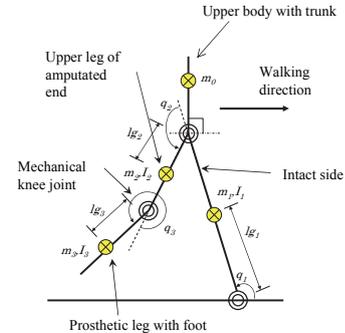

Fig. 1. Link segment model of the swing phase.

### C. Inertial Motion Generation Method

1) Via-point problem of differential equation
Definition: Via-point problem of differential equation

Let $x(t) \in \mathrm{R}^n$, $t \in [t_s, t_f]$ be a solution of the differential equation shown in Eq. (2) as follows:

$$\frac{d}{dt}x(t) = F(t, x(t)) .\qquad(2)$$

Finding the solution that satisfies $x(t_q) = x_q$, $t_q \in [t_s, t_f]$ is referred to as the via-point problem, where $x(t_q) = x_q$ is called the via-point.
Solution:

We define a representation of time using a reversal time $s$ as $f(s) = t_q - s$, where $t_q \in \mathbb{R}$ denotes the time when the solution is at the via-point. Note that $f(s)$ represents the time obtained by going back $s$[s] from $t_q$. Subsequently, $\tilde{x}(s) := x(f(s))$ is introduced. The differential equation (Eq. (2)) can now be rewritten as follows using the reversal time (Eq. (3)):

$$\frac{d}{ds}\tilde{x}(s) = -F(f(s), \tilde{x}(s)). \quad (3)$$

A solution of Eq. (2) $\tilde{x}(s) = x(t), t \in [t_s, t_q]$ that reaches $x(t_q)$ can be obtained by solving the initial value problem of Eq. (3) with the initial condition $\tilde{x}(0) = x(t_q)$. In addition, a solution $x(t), t \in [t_q, t_f]$ is obtained by solving the normal initial value problem of Eq. (2) with $x(t_q)$. By combining these two solutions, the solution of the via-point problem is obtained.

2) Motion generation method

A reference pattern for the joint motion $q$ is assumed to be given as $q^r(t) := [q_{12}^{rT}(t), q_3^r(t)]^T$, $q_{12}^r(t) := [q_1^r(t), q_2^r(t)]^T$, $t \in [t_0, t_f]$ in advance. In addition, the human motion at $q_1$ and $q_2$ is the same as the reference motion $q_1(t) = q_1^r(t), q_2(t) = q_2^r(t)$. Given this assumption, the inertial motion of the knee joint $q_3(t)$ is described in Eq. (4) by substituting $\tau_3 = 0$ and $q_{12}(t) = q_{12}^r(t)$ into Eq. (1) as follows:

$$\frac{d}{dt}\begin{bmatrix} q_3 \\ \dot{q}_3 \end{bmatrix} = \begin{bmatrix} \dot{q}_3 \\ -h_{33}^{-1}(q_{12}^r, q_3)\{h_{31}(q_{12}^r, q_3)\ddot{q}_1^r + h_{32}(q_{12}^r, q_3)\ddot{q}_2^r \\ + m_3 l_{g3}\{l_1 \sin(q_2^r + q_3)(\dot{q}_1^r)^2 + l_2 \sin(q_3)(\dot{q}_1^r + \dot{q}_2^r)^2\} \\ + m_3 l_{g3} \cos(q_1^r + q_2^r + q_3)g\} \end{bmatrix}, \quad (4)$$

where $l_i$, and $l_{gi}$ denote the length of each link and the distance of the center of gravity measured from the proximal end of each link, respectively; $m_i$ is the mass of the link; and $g$ stands for the acceleration due to gravity. Given the fact that Eq. (4) is a first-order differential equation of $[q_3(t), \dot{q}_3(t)]^T$, the desired inertial motion of the knee joint $q_3(t) = {}^{T_Q}q_3^{IM}(t)$ is generated by solving the via-point problem of Eq. (4) with the via-point $[q_3(T_Q), \dot{q}_3(T_Q)]^T = [q_3^r(T_Q), \dot{q}_3^r(T_Q)]^T$, where $T_Q$ is defined as follows:

$$T_Q = \arg\min_{t_Q} \int_{t_1}^{t_2} \{q_3^r(t) - {}^{t_Q}q_3^{IM}(t)\}^2 dt, \quad (5)$$

and $t_1, t_2 \in (t_s, t_f)$ are determined as the time at the maximum knee flexion angle and that at the maximum knee angular velocity, respectively [23].

In calculating Eq. (4), the inertial properties of the intact limbs are estimated through the method used in the literature [25], which is based on the data obtained from Japanese athletes. Table I lists the inertial characteristics of the prosthetic side obtained by assuming that a lightweight single axis prosthesis knee was used.

TABLE I. INERTIAL CHARACTERISTICS OF THE PROSTHESIS USED. $I_3$ DENOTES THE MOMENT OF INERTIA OF LINK 3 AROUND THE CENTER OF GRAVITY

| $m_3$ [kg] | $l_3$ [m] | $l_{g3}$ [m] | $I_3$ [kg·m²] |
|---|---|---|---|
| 1.0 | 0.501 | 0.425 | 0.0238 |

3) Result

Fig. 2 shows the inertial motion generated by the proposed method using the reference motion determined as $q^r(t) := [q_1^m(t), q_2^m(t), q_3^m(t)]^T$, where $q_j^m(j=1,2,3)$ denotes the joint angles measured in the experiments of the level walking of able-bodied individuals, as described in Section II-A.

Fig. 3 illustrates the error between the reference and the generated inertial motions for each phase of the initial swing, mid-swing, and terminal swing, as defined by Eq. (6):

$$\Delta q = \frac{1}{T_2 - T_1}\int_{T_1}^{T_2} |q_3^r(t) - {}^{t_Q}q_3^{IM}(t)|\,dt, \quad (6)$$

where $T_1$ and $T_2$ denote the start and end times of each phase, respectively. The initial swing starts at the toe-off and ends at the maximum knee joint flexion point. The mid-swing ends at the point corresponding to the maximum knee extension velocity. Finally, the terminal swing ends at the heel contact point of the prosthetic side.

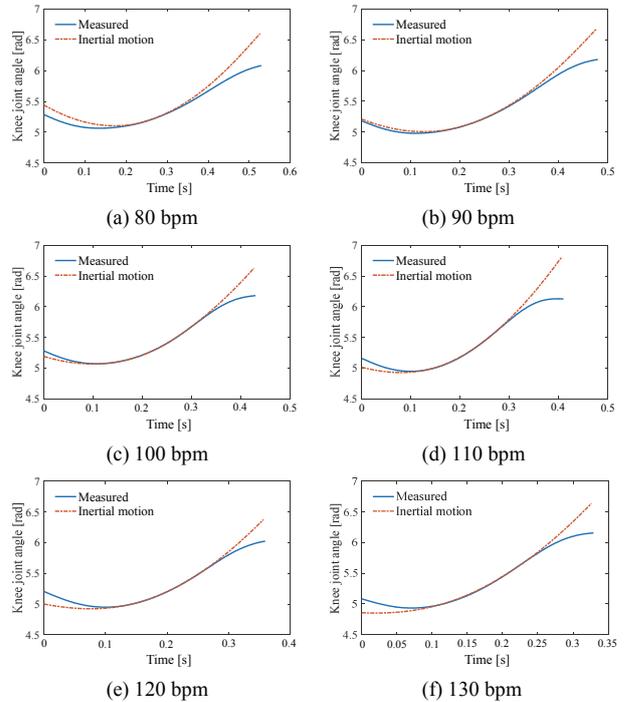

(a) 80 bpm   (b) 90 bpm
(c) 100 bpm   (d) 110 bpm
(e) 120 bpm   (f) 130 bpm

Fig. 2. Inertial motion generated by the proposed method. The measured knee angle is shown for comparison.

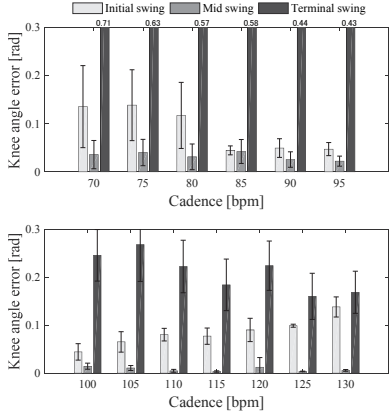

Fig. 3. Knee angle error between the generated and measured motions for each of the three phases.

These figures show that the inertial motion generated by the proposed method coincided with that measured in the mid-swing phase for a wide range of cadences (80–130 bpm).

*D. Kinematic Synergy in Prosthetic Gait with Inertial Motion*

1) Relationship between the intact joints and the prosthetic knee joint

The data matrix is defined in Eq. (7) as follows:

$$X := X_m - X_0, \quad (7)$$

$$X_m := [x_1, x_2, \cdots, x_n] \in R^{6 \times n}, \quad (8)$$

$$X_0 := [x_0, x_0, \cdots, x_0] \in R^{6 \times n}, \quad (9)$$

where $x_i := [q_1^m(t_i), \dot{q}_1^m(t_i), q_2^m(t_i), \dot{q}_2^m(t_i), q_3^{IM}(t_i), \dot{q}_3^{IM}(t_i)]$, $t_i := \{t \mid t_j < t_{j+1}, j = 1, \cdots, n-1\}$, and $x_0 := 1/n \sum_{i=1}^{n} x_i$. A reference motion of the first and second joints is defined as $q^r = [q_1^m, q_2^m]^T$, which is used for the generation of the inertial motion.

The singular value decomposition of matrix $X$ can be written as follows:

$$X = USV^T = \sum_{i=1}^{6} s_i u_i v_i^T, \quad (10)$$

where $U := [u_1, \cdots, u_6] \in R^{6 \times 6}$ and $V := [v_1 \cdots v_n] \in R^{n \times n}$ denote the left and right singular matrices, respectively, and $n$ denotes the data number in time. Matrix $S$ is defined as $S := \text{diag}[s_i] \ (i = 1, \cdots 6)$ and $s_j > s_{j+1}$, where $s_j$ is the singular value of matrix $X$.

The data matrix approximated using the largest $r$ singular value $s_1, \cdots, s_r$ can be defined as follows:

$$X_r := U_r S_r V_r^T = \sum_{i=1}^{r} s_i u_i v_i^T, \quad (11)$$

where $U_r := [u_1, \cdots, u_r]$, $V_r := [v_1, \cdots, v_r]$ and $S_r := \text{diag}[s_1, \cdots, s_r]$, where $r \leq 6$.

Vector $u_i$ represents the coordination between the motion of the joints, while vector $v_i$ denotes the temporal coordination representing the excitation pattern in time for each mode $u_i$.

There exists a coordination among joints, which is called kinematic synergy, if the total number of the modes $r$ that can sufficiently describe the original motion $X$ is less than the number of state variables[18].

2) Experimental results

The singular value decomposition in Eq. (11) was performed for the data collected during the level walking experiments described in Section II-A with cadences of 85, 100, 115, and 130 bpm. Note that this cadence range was chosen based on the analysis in Section III-B. Fig. 4 shows each element $u_{ij}$ of the column vector $u_i$ of the left singular matrix $U$ for each cadence. Fig. 5 presents vector $v_i$ of matrix $V$ for each cadence because the element corresponds to its changes in time.

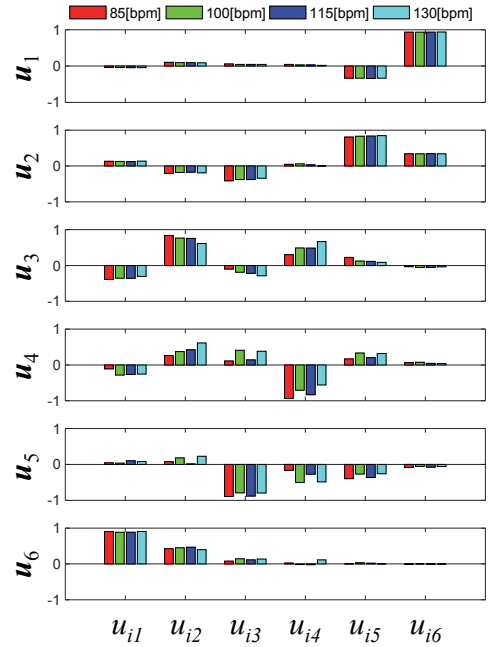

Fig. 4. Kinematic synergy.

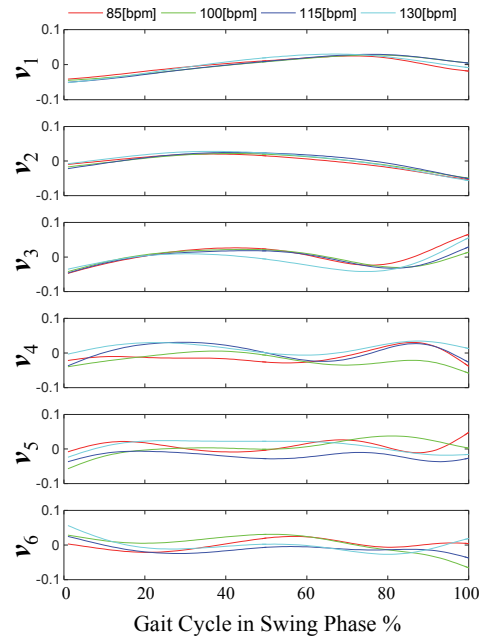

Fig. 5. Temporal coordination.

As shown in Table II, the cumulative contribution ratio exceeded 99%, with up to the fourth principal component being used for every cadence. These data indicate that kinematic synergy exists during level walking with a transfemoral prosthesis employing the inertial motion for a prosthetic knee joint, and that this synergy is present at a wide range of walking velocities.

The similar patterns of $u_i$ and $v_i$ among the cadences in Figs. 4 and 5 imply that similar synergies are present at a wide range of walking velocities from 85 to 130 bpm.

TABLE II. CONTRIBUTION RATIO

| $i$ | 85 bpm | 100 bpm | 115 bpm | 130 bpm |
|---|---|---|---|---|
| 1 | 91.67 | 93.58 | 94.94 | 93.90 |
| 2 | 8.08 | 6.19 | 4.93 | 5.90 |
| 3 | 0.15 | 0.14 | 0.10 | 0.10 |
| 4 | 0.05 | 0.06 | 0.01 | 0.06 |
| 5 | 0.02 | 0.01 | 0.001 | 0.01 |
| 6 | 0.001 | 0.0006 | 0.0002 | 0.0007 |

## III. METHOD FOR GENERATING THE KNEE JOINT MOTION

### A. Knee Joint Motion Generation Using Kinematic Synergy

1) Method

Based on the results of the synergy analysis in the previous section, Eq. (12) is satisfied given $r = 4$ in Eq. (11).

$$\begin{bmatrix} q_3^{IM}(t) \\ \dot{q}_3^{IM}(t) \end{bmatrix} = S_L U_r S_r (S_U U_r S_r)^+ \left\{ \begin{bmatrix} q_1^m(t) \\ \dot{q}_1^m(t) \\ q_2^m(t) \\ \dot{q}_2^m(t) \end{bmatrix} - S_U x_0 \right\} + S_L x_0, \quad (12)$$

$$S_U := [I_4, O_{4\times 2}] \in R^{4\times 6}, \quad S_L := [O_{2\times 4}, I_2] \in R^{2\times 6}. \quad (13)$$

Equation (12) can be rewritten as Eq. (14) by replacing $q_3^{IM}$ with $q_3^d$, which is the desired knee angle.

$$\theta_3^d = A\theta_{12}, \quad (14)$$

where $\theta_3^d := [q_3^d, \dot{q}_3^d]^T$ and $\theta_{12} := [q_1^m, \dot{q}_1^m, q_2^m, \dot{q}_2^m, 1]^T$. With Eq. (14), the desired knee angle and the velocity that can co-occur with the given intact joint angles can be calculated. Matrix $A$ is determined using the least squares method.

2) Result

Fig. 6 shows the knee joint trajectory estimated using the proposed equation (Eq. (14)) and the experimental data for each cadence. Coefficient matrix $A$ was determined for each cadence.

These results revealed that Eq. (14) can generate a knee motion that accords with its inertial motion by using the joint angles of the intact side for all the four cadences.

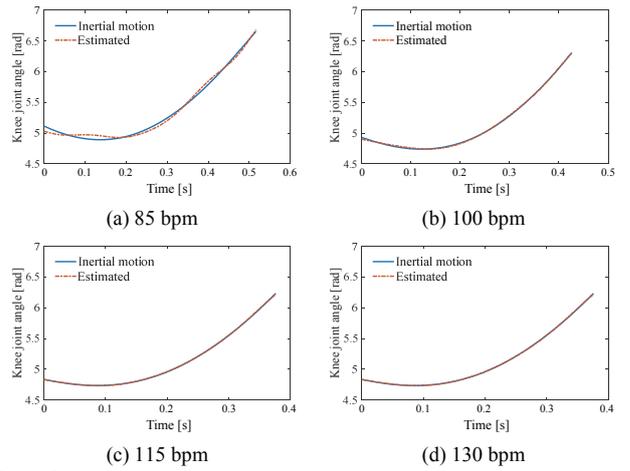

(a) 85 bpm (b) 100 bpm
(c) 115 bpm (d) 130 bpm

Fig. 6. Inertial motion and its trajectory estimated by the proposed method and calculated based on the joint angles of the intact side. Matrix $A$ in Eq. (14) is determined for each cadence based on the experimental results.

Fig. 7 shows the identified component of matrix $A$ for each cadence for one of the participants. Table III shows the coefficient of determinant $R^2$ of every element $a_{ij}$ for each of the six subjects (A through F). Statistical significances were found in most of the cases, which suggested that the components significantly depended on the walking velocity and appeared to change linearly with the cadence for several components.

TABLE III. COEFFICIENT OF DETERMINATION $R^2$

| | $a_{11}$ | $a_{12}$ | $a_{13}$ | $a_{14}$ | $a_{15}$ |
|---|---|---|---|---|---|
| | $a_{21}$ | $a_{22}$ | $a_{23}$ | $a_{24}$ | $a_{25}$ |
| A | 0.00001 | 0.28** | 0.03 | 0.001 | 0.03 |
| | 0.42** | 0.27** | 0.2** | 0.24** | 0.23** |
| B | 0.62* | 0.08* | 0.68* | 0.38** | 0.66** |
| | 0.19** | 0.07† | 0.3** | 0.12* | 0.24** |
| C | 0.0001 | 0.23** | 0.47** | 0.43** | 0.43** |
| | 0.63** | 0.002 | 0.62** | 0.56** | 0.64** |
| D | 0.54** | 0.1* | 0.64* | 0.69** | 0.6** |
| | 0.21** | 0.04† | 0.3** | 0.22** | 0.26** |
| E | 0.29** | 0.01 | 0.56** | 0.37** | 0.54** |
| | 0.09* | 0.34** | 0.15** | 0.25** | 0.07* |
| F | 0.11* | 0.06* | 0.1* | 0.13* | 0.13* |
| | 0.38** | 0.26** | 0.57** | 0.0001 | 0.51** |

\*\**p* < 0.01; \**p* < 0.05; †*p* < 0.1

### B. Generation Method of the Knee Joint Motion Adaptive to the Walking Velocity

Let us now consider a method of constructing the coefficient matrix $A$, such that it adapts to the walking velocity.

1) Relationship between the states of the intact limbs at the toe-off of the prosthetic side and the walking velocity

We hypothesize that Eq. (15) expresses the relationship between the cadence $v$ [bpm] and the states of the intact limbs $\boldsymbol{\theta}_{12}(t_{TO})$ as follows:

$$v = \boldsymbol{c}^T \boldsymbol{\theta}_{12}(t_{TO}), \quad (15)$$

where $\boldsymbol{c}$ denotes a coefficient vector.

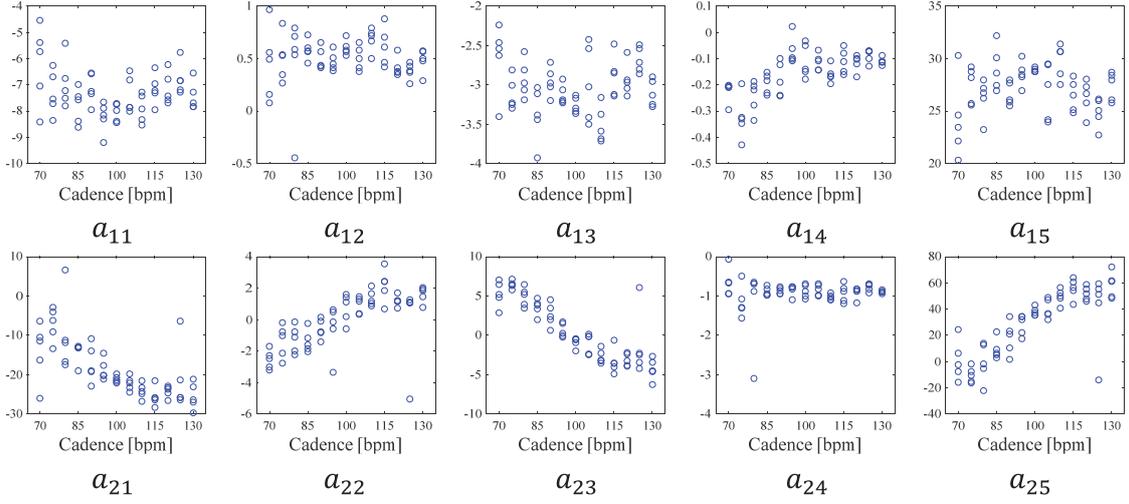

Fig. 7. Identified component of matrix $A$.

Fig. 8 shows the cadences estimated using Eq. (15) when the coefficient vector $\boldsymbol{c}$ was determined from the measured walking data using the least squares method. As can be seen from the figure, the relationship in Eq. (15) holds well.

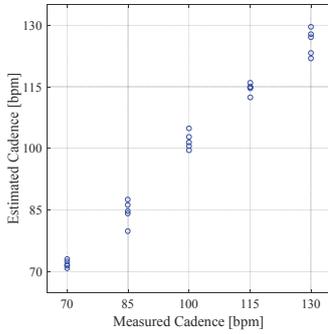

Fig. 8. Relationship between the measured cadences and those estimated from the toe-off motion.

2) Method for generating the prosthetic knee motion adaptive to the walking velocity

Based on the above results, we describe each element of matrix $A$ in Eq. (14) as a linear estimation using the states of the intact limbs at the toe-off of the prosthetic side $\boldsymbol{\theta}_{12}(t_{TO})$ as follows:

$$\boldsymbol{\theta}_3^d = A(\boldsymbol{\theta}_{12}(t_{TO}))\boldsymbol{\theta}_{12}, \quad (16)$$

$$A(\boldsymbol{\theta}_{12}(t_{TO})) := [a_{ij}] \in R^{2\times 5}, \quad (17)$$

$$a_{ij} = \boldsymbol{\beta}_{ij}^T \boldsymbol{\theta}_{12}(t_{TO}), \quad (18)$$

where $\boldsymbol{\beta}_{ij}$ denotes the coefficient vector for element $(i,j)$.

Matrix $A(\boldsymbol{\theta}_{12}(t_{TO}))$ was identified using the walking data collected at cadences of 85, 100, 115, and 130 bpm by the least squares method. Note that the cadence range used was determined based on our observation that the inclusion of data with 70, 75, and 80 cadences increases the errors between the knee joints generated by Eq. (16) and the inertial motions.

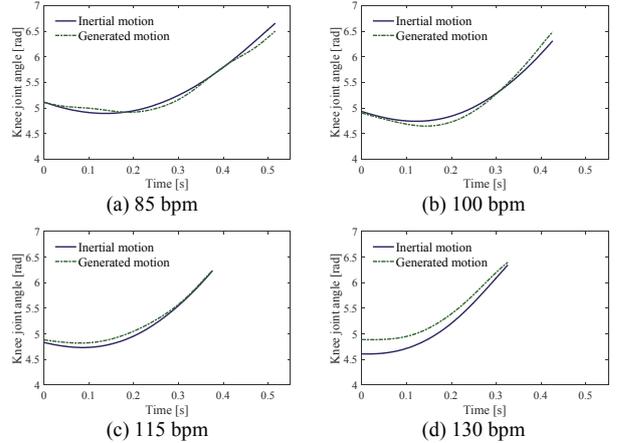

Fig. 9. Results of the knee joint motion generation by Eq. (16) adaptive to the walking velocity.

Fig. 9 shows examples of the inertial motion and the motion generated by the proposed method (Eq. (16)). Table IV lists the mean root-mean-square (RMS) error between them.

TABLE IV. RMS ERROR BETWEEN THE GENERATED KNEE JOINTS AND THE INERTIAL MOTION [rad]

| Cadence   | 85       | 100      | 115      | 130      |
|-----------|----------|----------|----------|----------|
| Mean (SD) | 0.00948  | 0.0847   | 0.0806   | 0.0880   |
|           | (0.0462) | (0.0262) | (0.0338) | (0.0374) |

## IV. SIMULATION EXPERIMENTS

### A. Dynamics and Control Input

Simulations were performed using the proposed knee motion generation method (Eq. (16)) to verify its effectiveness.

The dynamics of the lower limb are given by Eq. (19) as follows:

$$h_{31}(\boldsymbol{q}_{12}^m, q_3)\ddot{q}_1^m + h_{32}(\boldsymbol{q}_{12}^m, q_3)\ddot{q}_2^m + h_{33}(\boldsymbol{q}_{12}^m, q_3)\ddot{q}_3$$
$$+ m_3 l_{g3} \{l_1 \sin(q_2^m + q_3)(\dot{q}_1^m)^2 + l_2 \sin(q_3)(\dot{q}_1^m + \dot{q}_2^m)^2\}, \quad (19)$$
$$+ m_3 l_{g3} \cos(q_1^m + q_2^m + q_3)g - f_{damp} = \tau$$

where $f_{damp}$ denotes the damping force to prevent the extension of the knee beyond $2\pi$ and is defined by Eq. (20) as:

$$f_{damp} := K_f \left( \frac{1}{1 + e^{-300(2\pi - q_3)}} - 1 \right) \dot{q}_3. \quad (20)$$

The control input is designed as follows:
$$\tau = -K_p(q_3(t) - q_3^d(t)) - K_d(\dot{q}_3(t) - \dot{q}_3^d(t)), \quad (21)$$

where the desired motion $q_3^d(t), \dot{q}_3^d(t)$ is determined using the proposed method (Eq. (16)).

### B. Condition of the Simulation Experiments

The knee joint motion was simulated by numerically solving the initial value problem of Eq. (19) with the control input (Eq. (21)) for various conditions. The simulation experiments were performed under the three following conditions:

<u>Condition 1)</u> Cadence change: The measured motions of the intact limbs $\boldsymbol{q}_{12}^m$ for the cadences of 85, 100, 115, and 130 bpm recorded during the experiments described in Section II-A were used. The data from all the five trials for each cadence were used for the simulation.

<u>Condition 2)</u> Termination of walking: One of the subjects of the experiment in Section II-A participated in the new experiment. The subject was instructed to stop after walking at a self-selected velocity. The motions of the intact limbs in the final step measured in three trials during the experiments were used in the simulation.

<u>Condition 3)</u> Initiation of walking from the stopped state: The subject who participated in the experiment under condition 2 was instructed to walk at a self-selected velocity from the stopped state. The first step for each of the three trials was recorded.

Conditions 2 and 3 were set to examine the robustness of the proposed method with respect to the modulations in the walking velocity during a single swing phase. The knee angle was mechanically locked after the knee reached its full extension. The motions of the intact limbs $\boldsymbol{q}_{12}^m$ for these two cases were measured and used for the simulations. Note that the experimental setup to collect the data under simulation conditions 2 and 3 was the same as that mentioned in Section II-A.

### C. Results

1) Condition 1: Cadence change

Figs. 10–13 illustrate examples of the simulation results for condition 1 with cadences of 85, 100, 115, and 130 bpm, respectively. We found that the resultant knee motion was similar to the measured motion. Furthermore, the joint torque of the knee actuator was smaller during the mid-swing phase than that at the toe-off and before the heel contact. This result suggests that the knee motion in the mid-swing was close to its inertial motion, while a large torque was needed just after the toe-off to reach the inertial motion and just before the heel contact to converge to the final state. Please note that the zero torque observed just before the heel contact indicated that the knee joint reached its full extension and was mechanically locked.

Fig. 14 shows the minimum clearance between the heel position and the floor throughout the swing phase. The bars and error bars denote the means and standard deviations for the data for the six subjects, respectively. These results indicated that no collisions occurred, and that there was enough space to prevent the collisions between the foot and the floor for all the cadences during all the trials.

Fig. 15 illustrates the knee joint angle at the end of the simulation. The bars and error bars denote the means and standard deviations of the data from the six subjects, respectively. The figure demonstrates that the knee angle nearly reached a perfect knee extension ($2\pi$) for most of the conditions. This result indicated that the prosthetic knee joint was ready for the heel contact on the prosthetic side.

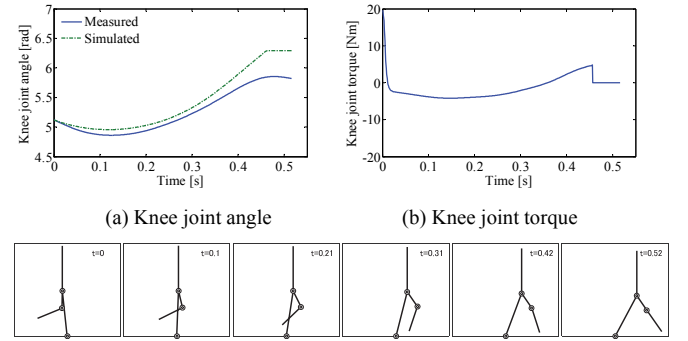

(a) Knee joint angle  (b) Knee joint torque

(c) Stick picture

Fig. 10. Simulation results (85 bpm under condition 1) using Eq. (16).

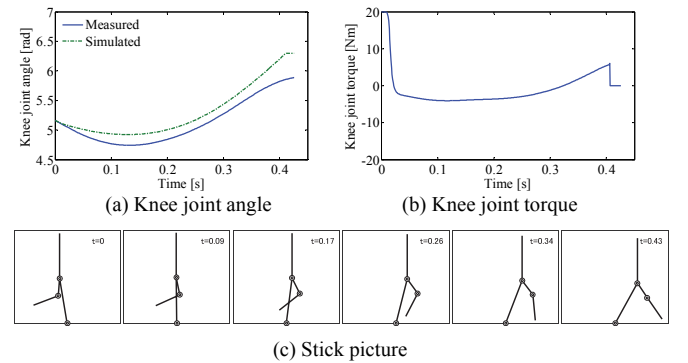

(a) Knee joint angle  (b) Knee joint torque

(c) Stick picture

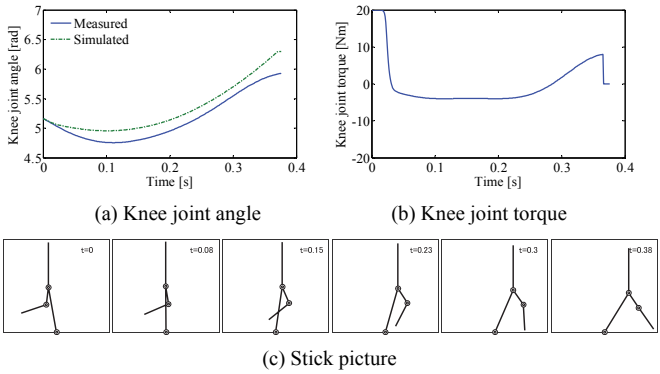

Fig. 11. Simulation results (100 bpm under condition 1) using Eq. (16).

(a) Knee joint angle   (b) Knee joint torque
(c) Stick picture
Fig. 12. Simulation results (115 bpm under condition 1) using Eq. (16).

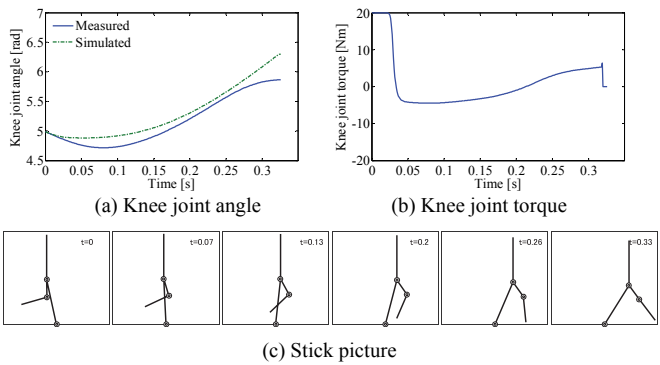

(a) Knee joint angle   (b) Knee joint torque
(c) Stick picture
Fig. 13. Simulation results (130 bpm under condition 1) using Eq. (16).

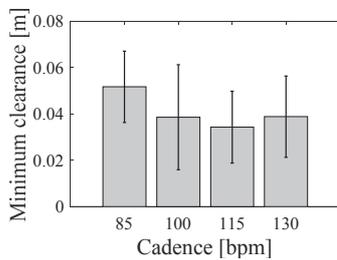

Fig. 14. Minimum clearance between the foot and the floor (condition 1).

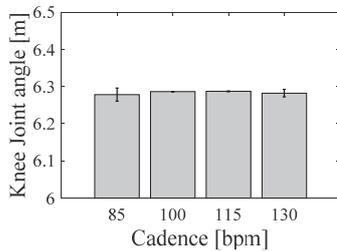

Fig. 15. Knee joint angle at the end of the simulation (or just before the heel contact (condition 1)).

2) Condition 2: Termination of walking

Fig. 16 shows an example of the simulation results for condition 2, that is, for the stopping case. We found that walking was successfully achieved without collision with the floor, and that a suitable degree of knee extension was achieved just before the heel contact even though there was a discrepancy between the measured and simulated knee motions. The resultant joint torque of the knee actuator was smaller during the swing phase than that just after the prosthetic toe-off and just before the heel contact.

The mean (and standard deviation) of the minimum clearance between the heel position and the floor throughout the swing phase was 0.0732[m] (0.00445). These results indicated that no collisions occurred between the foot and the floor during all the trials.

The mean (and the standard deviation) of the knee joint angle just before the heel contact was 6.28[rad] (0.00368). This result indicated that the prosthetic knee joint was ready for the heel contact on the prosthetic side.

3) Condition 3: Initiation of walking

Velocity adaptive method

Fig. 17 shows an example of the simulation results for condition 3, that is, for the "initiation of walking" case, in which the first step was made by the prosthetic side. We found that the simulated knee motion differed from the measured one, and the degree of the knee extension just before the heel contact was not enough to allow for the heel contact and could have resulted in a fall.

The mean (and standard deviation) of the minimum clearance between the heel position and the floor throughout the swing phase was 0.148[m] (0.00717). These results indicated that no collisions occurred between the foot and the floor during all the trials. The mean (and the standard deviation) of the knee joint angle just before the heel contact was 4.73[rad] (0.0877), which denoted that the prosthetic knee joint was not ready for the heel contact on the prosthetic side and resulted in a fall.

Thus, the motion generation method represented by Eq. (16), in which the knee motion was estimated or generated from the toe-off phase, may not be applicable in the "initiation of walking" case.

Regression for the walking initiation

Therefore, we assume that the kinematic synergy exists even in the "initiation of walking" case, and Eq. (14) is valid, but not Eq. (16). Accordingly, we identified matrix $A$ from the walking initiation data of a single trial. The desired knee joint angle was then generated using Eq. (14) along with the identified matrix $A$.

Fig. 18 illustrates the simulation results of the joint angle and the joint torque, as well as the stick pictures. These illustrations suggested that the kinematic synergy existed even during the initiation of walking.

The mean (and standard deviation) of the minimum clearance between the heel position and the floor throughout the swing phase was 0.0269[m] (0.00451), thereby indicating that no collisions occurred between the foot and the floor during all the trials.

The mean (and the standard deviation) of the knee joint angle just before the heel contact was 6.01[rad] (0.00807), which indicated that the prosthetic knee joint was ready for the heel contact on the prosthetic side.

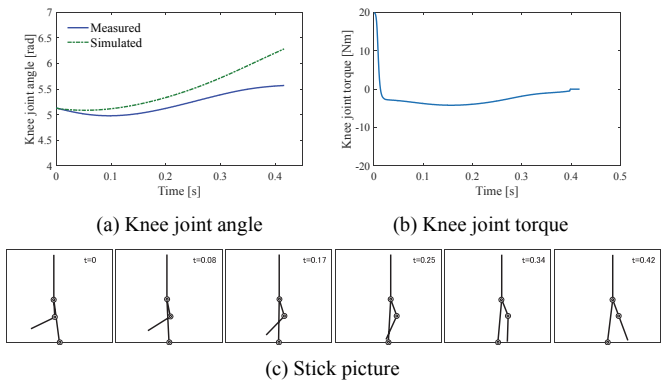

(a) Knee joint angle  (b) Knee joint torque

(c) Stick picture

Fig. 16. Simulation results (condition 2, termination of walking) using Eq. (16).

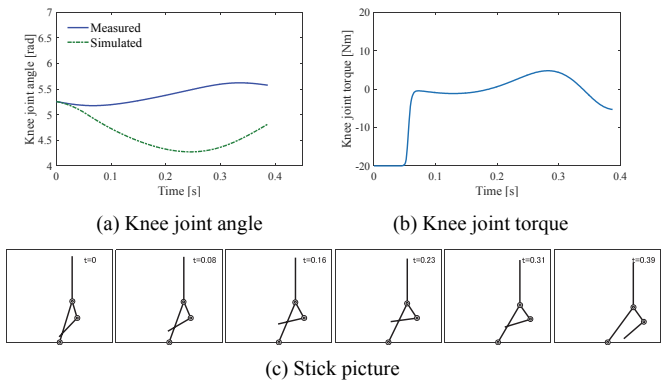

(a) Knee joint angle  (b) Knee joint torque

(c) Stick picture

Fig. 17. Simulation results (condition 3, initiation of walking) using Eq. (16).

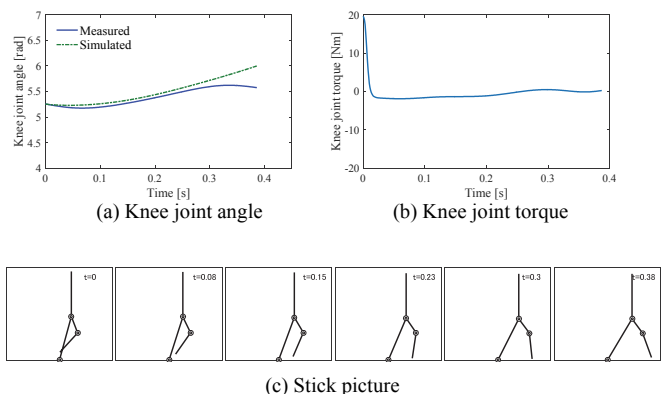

(a) Knee joint angle  (b) Knee joint torque

(c) Stick picture

Fig. 18. Simulation results (condition 3 with Eq. (14) using matrix $A$ calculated for the initiation of the walking condition).

## V. DISCUSSION AND CONCLUSIONS

The results of the inertial motion generation simulation showed that prosthetic users might be able to achieve a knee motion similar to that of able-bodied individuals in mid-swing using only the inertial motion although the weights of the prosthesis and the human body are greatly different from each other. The analysis of the kinematic synergy showed that the gait during level walking, by replacing the knee motion with the generated inertial motion, exhibits a synergy from 85 to 130 bpm cadence.

We developed a new motion generation method for the prosthetic knee joint based on these results by exploiting the inertial motion based on the linear regression for each walking velocity. This method has the advantage of generating a desired prosthetic knee motion using the motion of the intact limb in real time without using a timed trajectory. The regression coefficients used in the method were also revealed to depend on the walking velocity. Therefore, the motion generation method was updated to adapt to the walking velocity changes estimated from the motion of the intact limb at the prosthetic toe-off. The proposed method generated prosthetic knee motions that fit the reference gait of able-bodied individuals well.

In addition, the results of the feedback control simulations showed that the proposed method could generate a knee joint motion similar to that of able-bodied individuals for a wide range of cadences and situations of the termination of walking using only one formula and without requiring any changes to any of the coefficients. The analysis of the minimum clearance between the foot and the floor showed that no collision occurred. The result that the knee joint angles just before the heel contact almost reached their full extension demonstrated that the users were ready for the heel contact. These findings also suggested that the proposed method achieved safe walking without collision to the floor and falling at the heel contact. In addition, these results demonstrated the robustness of the proposed method to the walking velocity changes.

In contrast, the same method failed to generate feasible knee joint angles in the case of the initiation of walking from the stopped state. However, the original motion generation method proposed in the present paper could successfully generate a knee motion similar to that of able-bodied individuals during the initiation of walking. The analysis of the clearance to the floor and the knee extension just before the heel contact also showed its safety from the viewpoint of the clearance to the floor and the knee extension just before the heel contact. These were based on the fact that the kinematic synergy was observed even during the initiation of walking.

One of the contributions of the present study is the generation of the desired knee motion for a tracking control that is robust to the change and modulation of the walking velocity without a timed trajectory to achieve safe walking with a positive clearance to the floor and the knee extension before the heel contact. The conventional tracking control [9,10] uses the knee motion of able-bodied individuals, which is measured in advance, as the desired one. The echo control [15] utilizes the intact side's motion in the current step as the desired motion of the prosthesis in the next step. This method is not robust to the change in the walking speed and the velocity modulation, such as stopping. In the impedance control [6-8], the resultant knee angles could not be precisely controlled because they are generated by an exerted external force, including the hip moment and other disturbances. This might lead to an insufficient knee flexion during the mid-swing or extension before the heel contact that is important for safe walking. Another contribution was the leverage of the inertial motion for the desired motion based on the inertial motion generation technique as in [23]. The proposed method is expected to lead to an increase of the walking comfort because there exist studies showing that the inertial motion was used in the swing phase when the prosthesis users feel to walk easily [19,20]. Overall, the contribution of the present study is the achievement of a safe and natural knee motion leveraging the inertial motion from the initiation of walking through walking at a constant velocity in a steady state to the termination of walking without a timed trajectory, which were all seamlessly connected.

Employing the inertial motion could increase energy efficiency, which could contribute to the reduction of the weight of the actuator and the battery. The inertial motion generation method is expected to be robust to the differences in the inertial property from the fact that the weight of the prosthesis in the present research was much lighter than that of the able-bodied individual's body, and the able-bodied individuals also used the inertial motion in mid-swing [23].

In the inertial motion calculation, only one inertial parameter set was used to examine the effectiveness of our proposed method, which is one of the limitations of this study. The present research showed that the prosthetic knee trajectory similar to that in able-bodied individuals can also be expressed using the inertial motion of the prosthesis although it has a lighter weight. Thus, a similar tendency can be expected for the prostheses with a weight ranging between that of the prosthesis given in the paper and that in able-bodied individuals. It also shows the robustness to the difference of the inertial property.

In the study, 85 to 130 bpm walking was mainly dealt with because large errors were found in the velocity adaptive method in lower velocities (e.g., 70 to 80 bpm). This is also one of the limitations of the proposed method. The mean cadence at the preferred walking velocity in a non-amputated human is known to be approximately 120 bpm [26,27]. As regards the cadence in the preferred walking velocity with the transfemoral prosthesis, one study reported that the mean cadences with non-microprocessor- and microprocessor-controlled prostheses were approximately 100 bpm [28]. However, it depends on the physical conditions of the users, the prosthesis types, and the experience of prosthesis use. The 85–130 cadence range, where the present research focuses on, is thought to cover this range. However, the cadence of in-house ambulation could be smaller than these values. The proposed method is expected to be used even in the lower cadence range because it could be used even in the initiation of walking. However, the effect of the proposed method in the lower cadence range, including 70–80 bpm, should be investigated as a future work. In addition, the intact individuals' knee joint motions were utilized as the reference of the desired motion. Using the prosthetic users' own knee motion as the reference is impossible and is another limitation of the proposed method. Some countermeasures for this exist, including the selection of the most suitable gait from the gait database measured from different persons. The method of choosing the most suitable patterns is an important future work.

Future studies will include the implementation of the proposed method in an actual prosthesis and an evaluation of the method from a biomechanics perspective. The robustness of the proposed method to the changes in the input signals for a wider variety of signals, such as a larger hip joint motion, should be investigated. Investigating the ease of walking with the proposed method is also an important future work to validate our hypothesis that leveraging the inertial motion in the prosthetic knee control would increase the ease of walking.

**Hiroshi Sano** was born in Kizugawa City, Kyoto, Japan in 1991. He received his B.S. and M.S. degrees in Engineering from Ritsumeikan University in 2014 and 2016, respectively. In 2016, he joined Canon IT Solutions Inc. Between 2013 and 2015, when he was a member of a human robotics laboratory, he studied the comfort aspects of the motion control of transfemoral prostheses by leveraging their inertial motion.

**Takahiro Wada (M'99)** received his B.S. degree in Mechanical Engineering, his M.S. degree in Information Science and Systems Engineering, and his Ph.D. degree in Robotics from Ritsumeikan University, Japan in 1994, 1996, and 1999, respectively. In 1999, he became an assistant professor at Ritsumeikan University. In 2000, he joined Kagawa University in Takamatsu, Japan as an assistant professor in the Department of Intelligent Mechanical Systems Engineering, Faculty of Engineering. He was promoted to associate professor in 2003. He has been a full-time professor at Ritsumeikan University in the College of Information Science and Engineering since 2012. In 2006 and 2007, he spent half a year at the University of Michigan Transportation Research Institute, Ann Arbor as a visiting researcher.

His current research interests include robotics, human machine systems, and human modeling. He is also interested in rehabilitation robotics, automotive safety by driver assistance systems, and various other areas of human–machine physical interactions.

Dr. Wada is a member of IEEE (EMBS, RAS, SMC, and ITSS), the Human Factors and Ergonomics Society (HFES), the Robotics Society of Japan (RSJ), the Society of Automotive Engineers of Japan (JSAE), the Society of Instrument and Control Engineers (SICE), and the Japan Society of Mechanical Engineers (JSME). He received the Best Paper Award from JSAE in 2008 and 2011 and the Outstanding Oral Presentation from the Society for Automotive Engineers (SAE) in 2010, among other awards.